\documentclass[conference]{IEEEtran}
\IEEEoverridecommandlockouts

\usepackage{cite}
\usepackage{amsmath,amssymb,amsfonts}
\usepackage{algorithm}
\usepackage{algpseudocode}
\usepackage{amsmath}
\usepackage{graphicx}
\usepackage{algpseudocode}
\usepackage{graphicx}
\usepackage{textcomp}
\usepackage{xcolor}
\usepackage{booktabs} 
\usepackage{graphicx} 
\usepackage{tabularx} 
\usepackage{hyperref}  
\def\BibTeX{{\rm B\kern-.05em{\sc i\kern-.025em b}\kern-.08em
    T\kern-.1667em\lower.7ex\hbox{E}\kern-.125emX}}

\begin{document}

\title{Modality-Invariant Bidirectional Temporal Representation Distillation Network for Missing Multimodal Sentiment Analysis}

\author{\IEEEauthorblockN{Xincheng Wang, Liejun Wang$^{\dagger}$, Yinfeng Yu$^{\dagger}$, Xinxin Jiao}
\thanks{$^{\dagger}$Both Liejun Wang and Yinfeng Yu are corresponding authors.}
\IEEEauthorblockA{\textit{School of Computer Science and Technology} 
\textit{, Xinjiang University,}
\textit{Urumqi, China }\\
Email: wljxju@xju.edu.cn, yuyinfeng@xju.edu.cn}}

\maketitle

\begin{abstract}
Multimodal Sentiment Analysis (MSA) integrates diverse modalities—text, audio, and video—to comprehensively analyze and understand individuals' emotional states. However, the real-world prevalence of incomplete data poses significant challenges to MSA, mainly due to the randomness of modality missing. Moreover, the heterogeneity issue in multimodal data has yet to be effectively addressed. To tackle these challenges, we introduce the Modality-Invariant Bidirectional Temporal Representation Distillation Network (MITR-DNet) for Missing Multimodal Sentiment Analysis. MITR-DNet employs a distillation approach, wherein a complete modality teacher model guides a missing modality student model, ensuring robustness in the presence of modality missing. Simultaneously, we developed the Modality-Invariant Bidirectional Temporal Representation Learning Module (MIB-TRL) to mitigate heterogeneity. 
\end{abstract}


\section{Introduction}
\begin{figure}[t]
	\centering
	\includegraphics[width=\linewidth]{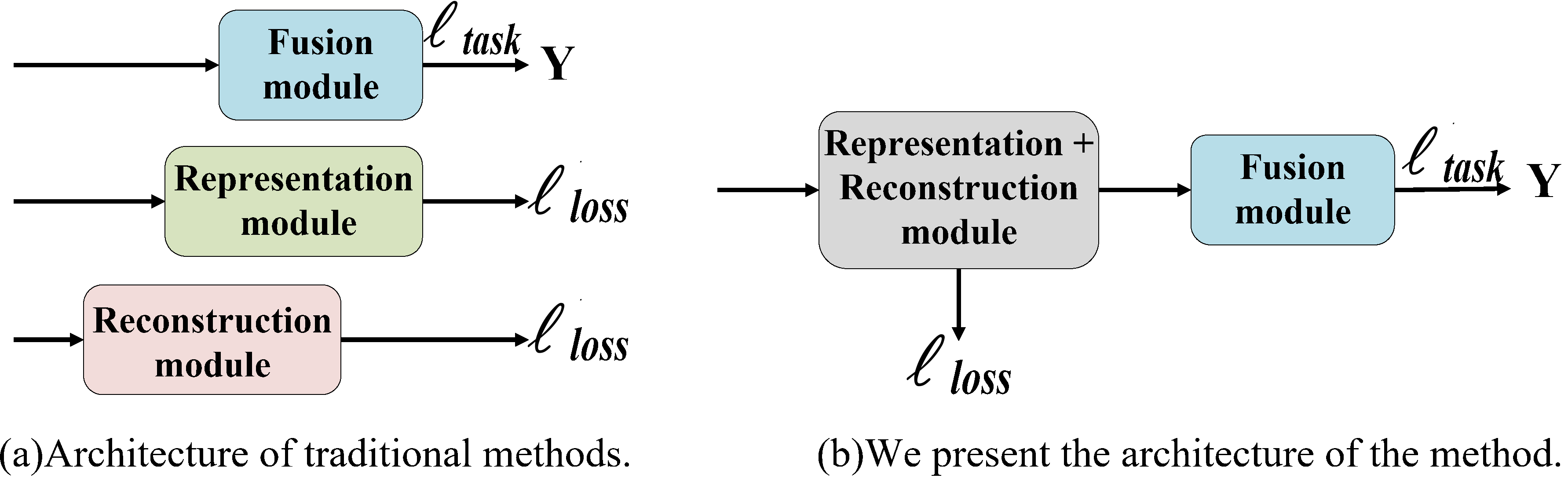} 
	\caption{Traditional framework vs. our proposed framework.}
	\label{1}
\end{figure}

Multimodal Sentiment Analysis (MSA) \cite{1,2,3} utilizes various information modalities such as text, audio, and video to analyze and understand individuals' affective states and intentions. MSA is widely applied in HCI systems \cite{41,42,43,44,45}, mental health \cite{46}, and other fields. Studies have shown that MSA provides more nuanced and accurate emotional assessments than single-modality analysis \cite{4,5}. However, many existing studies\cite{6,7,8,9,10} focus on scenarios where all modalities are present or one is entirely missing, failing to capture the more common real-world scenario of random modality absence. This oversight limits the practical applicability of these models. To address this gap, we propose a novel approach that models the randomness of modality missing. We employ knowledge distillation techniques, using a complete modality network as the “teacher” and a network with random modality missing as the “student.”  In addition, in the presence of modality random missing, multimodal data still faces the challenge of modal heterogeneity, which poses a significant challenge for multimodal fusion and improving accuracy \cite{11}\cite{12}. Modality representation learning has been an effective coping strategy in previous MSA studies. For example, in \cite{13,14,15}, researchers mitigated modal heterogeneity by decoupling multimodal features into modality-invariant and modality-specific features and using loss constraints. However, most of the above methods were proposed assuming no missing modalities and similar approaches have been applied less in the study of MSA with missing modalities. Although a modality-invariant representation learning module was designed in\cite{16} to mitigate inter-modal heterogeneity, directly using these representations may result in incomplete information when a modality is missing. Modality representation learning and modality missing reconstruction usually exist as independent modules in existing studies of missing modality MSA \cite{16,17,18}. The reconstruction module employs reconstruction loss, while the representation learning module applies similarity or difference loss constraints. This split design requires additional network structures to predict outputs, increasing complexity and redundancy, as shown in Fig.~\ref{1}.

\begin{figure*}[t]
	\centering
	\includegraphics[width=\textwidth, height=8cm]{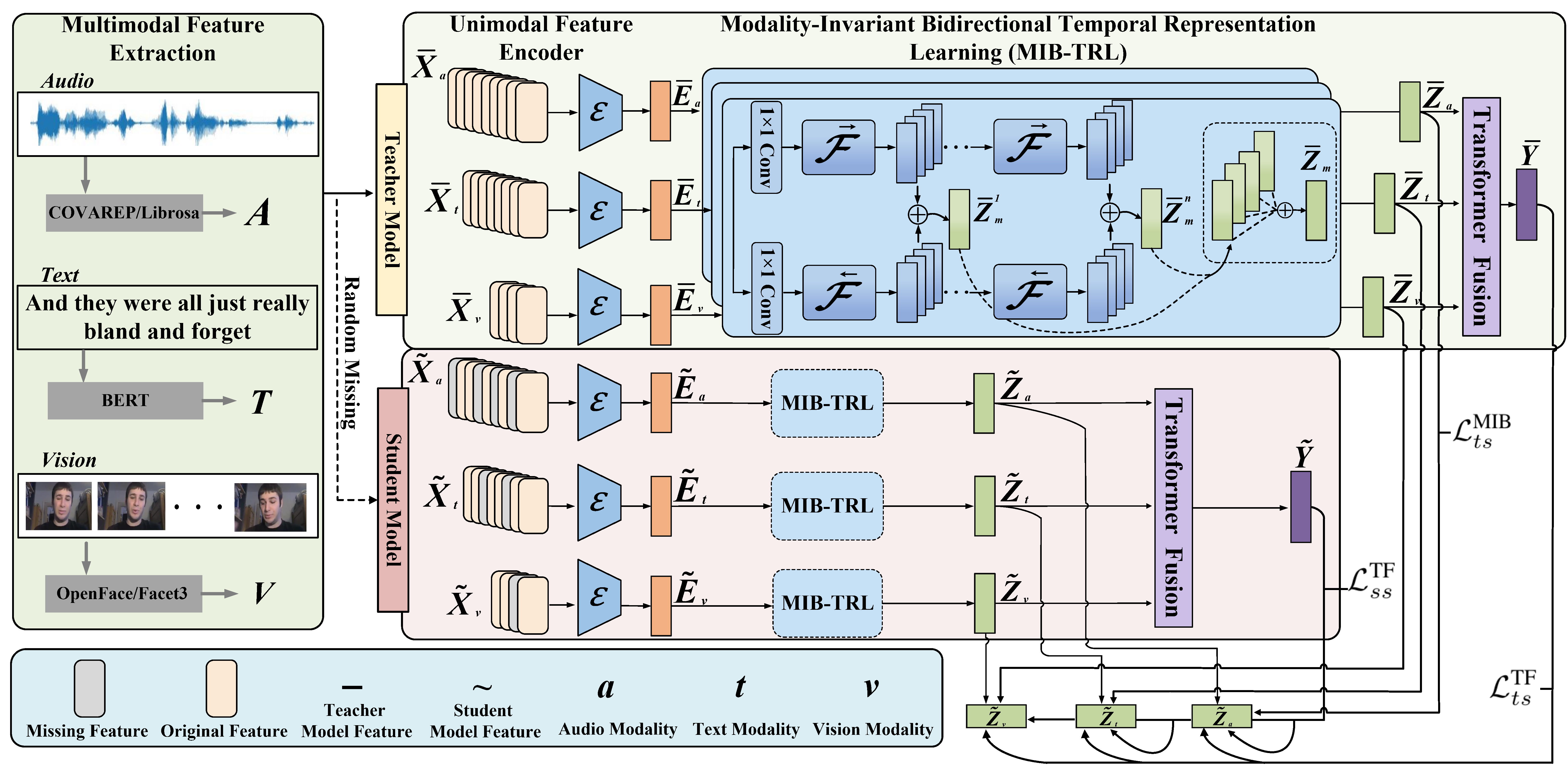} 
	\caption{Present the overall framework of the MITR-DNet methodology.}
	\label{2}
\end{figure*}
To this end, we propose, for the first time, integrating modality reconstruction and modality representation learning into a single module called Modality-Invariant Bidirectional Temporal Representation Learning (MIB-TRL). MIB-TRL achieves feature alignment and mitigates feature heterogeneity by projecting the generated features into the same feature subspace. It also provides a solid foundation for subsequent multimodal feature fusion. In MIB-TRL, we focus solely on the modality-invariant subspace, which ensures consistent modality information and further simplifies the multimodal architecture.
\section{Methodologies}
	We consider three modalities: audio ($a$), text ($t$), and vision ($v$). The complete input feature sequence is defined as $\bar{X}_m \in \mathbb{R}^{T_m \times f_m}$, where $m$ represents the three modalities, i.e., $m \in \{a, t, v\}$, with $T_m$ and $f_m$ denoting the time steps and feature dimensions, respectively. To simulate missing modality features in real-world scenarios, we introduce a masking function $F(\cdot)$ and a randomly generated time mask $g_{m} \in \left \{ 0,1 \right \} ^{T_{m} } $, resulting in the incomplete sequence $\tilde{X}_m = F(\bar{X}_m, g_m)$. As shown in {Fig.~\ref{2}}, different tools are employed for feature extraction from each modality. In unimodal feature encoder \(\mathcal{E}\), we extract long-term contextual features $E_m$ using self-attentive encoders for $a$, $t$ and $v$ modalities. Both settings are applicable when the diacritical mark $- / \sim $ is not used.

\subsection{MIB-TRL module}
\begin{table*}[h]
	\centering
	\caption{Results of experiments on CMU-MOSI and CH-SIMS datasets in both complete and random modality missing settings.}
	\label{tab1}
	\small 
	\renewcommand{\arraystretch}{1}
	\resizebox{\textwidth}{!}{ 
		\begin{tabular}{@{}lcccccp{0.5cm}ccccccc@{}}
			\specialrule{1.1pt}{0pt}{0pt} 
			& \multicolumn{5}{c}{\textbf{CMU-MOSI (Complete Modality)}} & \phantom{abc} & \multicolumn{5}{c}{\textbf{CH-SIMS (Complete Modality)}} \\
			\cmidrule{2-6} \cmidrule{8-12}
			\textbf{Models} & \textbf{MAE ($\downarrow$)} & \textbf{Corr ($\uparrow$)} & \textbf{Acc-5 ($\uparrow$)} & \textbf{Acc-2 ($\uparrow$)} & \textbf{F1 ($\uparrow$)} && \textbf{MAE ($\downarrow$)} & \textbf{Corr ($\uparrow$)} & \textbf{Acc-3 ($\uparrow$)} & \textbf{Acc-2 ($\uparrow$)} & \textbf{F1 ($\uparrow$)} \\
			\midrule
			TFN\cite{6} & 0.942 & 0.662 & 40.92 & 77.11/78.56 & 77.09/78.60 & & 0.438 & 0.577 & 65.50 & 75.05 & 75.31 \\
			MulT\cite{7} & 0.874 & 0.703 & 44.07 & 79.54/80.74 & 79.42/80.68 & & 0.443 & 0.558 & \textbf{66.08} & 77.75 & 77.69 \\
			TFR-Net\cite{18} & 0.766 & 0.779 & 50.00 & 82.75/84.71 & 82.69/84.70 & & 0.441 & 0.564 & 63.38 & 78.99 & 78.14 \\
			EMT-DLFR\cite{17} & 0.708 & 0.796 & 53.84 & 82.36/84.25 & 82.29/84.23 & & \textbf{0.403} & 0.610 & 65.65 & 78.70 & 78.86 \\
			\textbf{MITR-DNet} & \textbf{0.708} & \textbf{0.797} & \textbf{54.33} & \textbf{83.28/84.91} & \textbf{83.22/84.90} & & 0.411 & \textbf{0.617} & 64.99 & \textbf{81.18} & \textbf{81.05} \\
			\midrule \midrule
			& \multicolumn{5}{c}{\textbf{CMU-MOSI (Incomplete Modality)}} & \phantom{abc} & \multicolumn{5}{c}{\textbf{CH-SIMS (Incomplete Modality)}} \\
			\cmidrule{2-6} \cmidrule{8-12}
			TFN\cite{6}    & 1.311 & 0.312 & 23.6 & 60.0/59.6& 56.6/56.4 && 0.234& 0.257 &30.4& 37.3 & 37.2 \\
			MulT\cite{7}  & 1.274 & 0.330 & 24.6 &63.3/63.5 & 61.9/62.3 && 0.240 & 0.242& 30.3& 37.6 & 36.9 \\
            MISA\cite{8}  & 1.119 & 0.427 & 27.8 &64.7/64.6 & 60.0/60.0 && 0.295 & 0.030& 26.8& 34.7 & 28.4 \\
			TFR-Net\cite{18} & 1.175 & 0.451 &29.4 & 67.2/67.4 & 65.5/65.9 && 0.240 & 0.240 &29.1 & 37.4 & 37.0 \\
			EMT-DLFR\cite{17}  & 1.111 & \textbf{0.487} & \textbf{35.0} & 69.5/70.0 & 69.2/69.9 && 0.217& 0.283 & 31.6 & 37.9 & 38.0 \\
			\textbf{MITR-DNet}  &\textbf{1.105} & 0.479 & 34.4& \textbf{69.6/70.3} & \textbf{69.4/70.2}&& \textbf{0.216} & \textbf{0.294}& \textbf{32.0} & \textbf{39.3} & \textbf{39.2} \\
			\specialrule{1.2pt}{0pt}{0pt} 
		\end{tabular}
	}
\end{table*}
\textbf{Time-aware generation module $\mathcal{F} $.}  In missing MSA processing, accurate reconstruction of missing features is crucial for sentiment prediction. Inspired by the WaveNet approach to speech generation \cite{20}, we apply the Dilated Causal Convolution technique to the time-aware generation module $\mathcal{F}$. This module progressively predicts and reconstructs missing modality features using an autoregressive approach, ensuring that each step relies solely on the available inputs, thereby avoiding future information interference. For each layer \(i \in \{1, 2, \dots, n\}\), perform the following operations:
\begin{eqnarray}
y_{dc} = \text{DConv}(y, d_i), \quad \xi = \tanh(y_{dc}) \times \sigma(y_{dc}).
\end{eqnarray}
where \(d_i = 2^{i-1}\) is the dilation rate and DConv is the dilated causal convolution. The output $y'$ of module $\mathcal{F}$ is then obtained by the following operation:
\begin{eqnarray}
\eta = \text{Conv}_{1\times1}(\xi), \quad y' = \eta + y.
\end{eqnarray}
\textbf{MIB-TRL module.} {Fig.~\ref{2}} shows the MIB-TRL module designed for each input modality. The method first processes the input data through two identical convolutional layers and then splits the data stream into two directions. Each direction consists of n modules $\mathcal{F} $.  In both directions, the output of the $\mathcal{F} $ module at the same $i$ position is integrated to obtain $Z_{m}^{i} $. Finally, all $Z_{m}^{i} $ are fused to obtain the full long-term contextual representation of $m$ modality.

\newcommand{\Input}{\item[\textbf{Input:}]}
\newcommand{\Output}{\item[\textbf{Output:}]}
\begin{eqnarray}
	Z^i_m = f^i_m(\overrightarrow{\mathcal{F} } ) + f^i_m(\overleftarrow{\mathcal{F} } ) ,        Z_{m} = {\textstyle \sum_{i=1}^{n}} Z_{m}^{i} .
\end{eqnarray}
 
\subsection{Transformer Fusion (TF)}
Previous studies \cite{21}\cite{22} have shown that the text modality contains more accurate semantic information and plays a crucial role in sentiment prediction. In this paper, our innovative Transformer Fusion structure fully leverages the central role of the text modality.
\small
\textbf{Visual-to-Audio auxiliary:$v\leftrightarrow t\leftrightarrow a$}
\begin{eqnarray}
	\mathcal{SA}_{t\leftrightarrow v} = \mathcal{SA} (X_t + X_v) + (X_t + X_v),
\end{eqnarray}
\vspace{-20pt} 
\begin{eqnarray}
	Y_{t\leftrightarrow v} = MLP (\mathcal{SA}_{t\leftrightarrow v}), \quad Q_{t\leftrightarrow v} = \mathcal{SA}_{t\leftrightarrow v},
\end{eqnarray}
\vspace{-20pt} 
\begin{eqnarray}
	V_{t\leftrightarrow a} = K_{t\leftrightarrow a} = \mathcal{SA}_{t\leftrightarrow a},
\end{eqnarray}
\vspace{-20pt} 
\begin{eqnarray}
	Y_{v\leftrightarrow t\leftrightarrow a} = MLP \left( \mathcal{CA}(Q_{t\leftrightarrow v}, K_{t\leftrightarrow a}, V_{t\leftrightarrow a}) \right),
\end{eqnarray}
\vspace{-20pt} 
\begin{eqnarray}
	{Y}' _{v\leftrightarrow t\leftrightarrow a} = MLP \left( Y_{t\leftrightarrow v} \right) + Y_{v\leftrightarrow t\leftrightarrow a} + MLP \left( Y_{t\leftrightarrow v} \right).
\end{eqnarray}
\textbf{Audio-to-Visual auxiliary: $a\leftrightarrow t\leftrightarrow v$}
\begin{eqnarray}
	{Y}' _{a\leftrightarrow t\leftrightarrow v} = MLP (Y_{t\leftrightarrow a}) + Y_{a\leftrightarrow t\leftrightarrow v} + MLP (Y_{t\leftrightarrow a}).
\end{eqnarray}
$\mathcal{SA}(X)$ denotes self-attention and $\mathcal{CA}(X)$ denotes cross-modal attention. MLP denotes multilayer fully connected. The final overall transformer output is:
\begin{eqnarray}\label{17}
	Y = {Y}' _{v\leftrightarrow t\leftrightarrow a} + 	{Y}' _{a\leftrightarrow t\leftrightarrow v}.
\end{eqnarray}
\normalsize

\subsection{Training objective}
This loss function combines the regression task loss, distillation loss, reconstruction loss, and SimSiam loss\cite{23}. When it is a complete modality setting, there is no need to perform the distillation operation, and the teacher model accomplishes the task alone with an overall loss function of:
\begin{equation}
	\mathcal{L} = |Y_{label} - \bar{Y} |.
\end{equation}
When it is an incomplete modality setting:
\begin{equation}
	\mathcal{L} = |Y_{label} - \tilde{Y} |+ \mathcal{L}_{\text{$dis$}} + \mathcal{L}_{\text{$rec$}} + \mathcal{L}_{\text{$sim$}.}
\end{equation}
\textbf{Distillation loss:} This loss comprises the distillation loss between the teacher and student models after the MIB-TRL and TF modules, along with the self-distillation loss of the student model.
\small
\footnotesize
\begin{equation}
	\mathcal{L}_{\text{$ts$}}^{\text{MIB}} =  {\textstyle \sum_{m}^{\left \{ a,t,v \right \} }} \left \| \bar{Z}_{m}  -\tilde{Z}_{m}  \right \|^{2},\mathcal{L}_{\text{ts}}^{\text{TF}} = {\textstyle \sum_{m}^{\left \{ a,t,v \right \} }}   \left \|\bar{Y} -\tilde{Z}_{m}   \right \| ^{1}.
\end{equation}
\normalsize

\small
\begin{equation}
	\mathcal{L}_{\text{ss}}^{\text{TF}} = {\textstyle \sum_{m}^{\left \{ a,t,v \right \} }} \left \| \tilde{Y} - \tilde{Z}_{m} \right \|^{2} .
\end{equation}
\normalsize
The total distillation loss is:
\begin{equation}
	\mathcal{L}_{\text{$dis$}} = \lambda_1 (\mathcal{L}_{\text{$ts$}}^{\text{MIB}} + \mathcal{L}_{\text{$ts$}}^{\text{TF}} + \mathcal{L}_{\text{$ss$}}^{\text{TF}}).
\end{equation}
\textbf{Reconstruction loss:} After the student model processes the data through the unimodal feature encoder  \(\mathcal{E}\) and MIB-TRL module, the original feature $X \in \{A, T, V\}$ is referred to, and a Multi-Layer Perceptron (MLP) decoder is used to reconstruct the complete modality sequence. Reconstruction losses $\mathcal{L}_{\text{$rec$}}^{\text{en}}$ and $\mathcal{L}_{\text{$rec$}}^{\text{MIB}}$ are obtained respectively. The reconstruction loss can be expressed as:
\small
\begin{equation}
	\mathcal{L}_{\text{rec}} =  {\textstyle \sum_{m}^{\left \{ a,t,v \right \} }} \text{smooth}_{L1} \left( X - \text{MLP}(\tilde{E}_m \text{ or } \tilde{Z}_m) \cdot (1 - M_m) \right).
\end{equation}
\normalsize
where $\tilde{E}_m$ and $\tilde{Z}_m$ represent the features processed by the unimodal feature encoder and MIB-TRL module, respectively, $M_{m} $ denotes the time mask that excludes losses at unmasked locations. The total reconstruction loss is expressed as:
\begin{equation}
	\mathcal{L}_{\text{rec}} = \lambda_2 \left( \mathcal{L}_{\text{rec}}^{\text{en}} \right) + \mathcal{L}_{\text{rec}}^{\text{MIB}}.
\end{equation}
\textbf{Simsiam loss:}  We impose loss constraints on the available features to avoid increasing network complexity, but the issue of collapsing solutions in representation learning persists \cite{23}. To address this, we introduce SimSiam loss\cite{23}, which prevents collapse by employing a stop-gradient mechanism, ensuring more stable multimodal feature representations.
\begin{equation}
	\begin{split}
		\mathcal{L}_{\text{sim}} = \frac{1}{2}\mathcal{D} \big(f(\tilde{Z}_m \text{ or } \tilde{E}_m), \text{sg}(h(\bar{Z}_m \text{ or } \bar{E}_m))\big) + \\
		\frac{1}{2}\mathcal{D} \big(f(\bar{Z}_m \text{ or } \bar{E}_m), \text{sg}(h(\tilde{Z}_m \text{ or } \tilde{E}_m))\big).
	\end{split}
\end{equation}
Where $f$ and $h$ are MLP with different mapping dimensions. $sg$ denotes stopping the gradient operation. $\mathcal{D}$ denotes negative cosine similarity minimization on the features($p_1, z_2)$:
\begin{equation}
	\mathcal{D}(p_1, z_2) = -\frac{p_1 \cdot z_2}{\|p_1\|_2 \cdot \|z_2\|_2}.
\end{equation}
The total SimSiam loss is:
\begin{equation}
	\mathcal{L}_{\text{sim}} = \mathcal{L}_{\text{sim}}^{\text{MIB}} + \lambda_3 (\mathcal{L}_{\text{sim}}^{\text{en}}).
\end{equation}

\section{Experimental Setup}
\subsection{Datasets}
The CMU-MOSI dataset\cite{24} examines sentiment in English-language videos, with intensity scores spanning from -3 (strongly negative) to 3 (strongly positive). The CH-SIMS dataset\cite{25} comprises 2281 video clips from 60 Chinese videos sourced from movies and variety shows, with sentiment intensities from -1 (strongly negative) to 1 (strongly positive).

\subsection{Implementation details}\label{SCM}
Our model is implemented using PyTorch and optimized with Adam, with a batch size of 32 and an early stopping set at eight epochs. We initialized three seeds to ensure reproducibility, averaging the results for consistency. The hyperparameters ${\lambda}_1, {\lambda}_2, {\lambda}_3$ were optimized using a grid search approach. For the MOSI dataset, the optimal values were set to 0.01, 0.1, and 0.1, respectively, while for the SIMS dataset, the optimal values were determined to be 0.3, 0, and 0, respectively.
\begin{table}[htbp]
	\centering
	\caption{The number $i$ of ablations $\mathcal{F}$ on the CH-SIMS dataset. }
	\label{22}
	\setlength{\tabcolsep}{6pt} 
	\begin{tabular}{@{}cccccc@{}}
		\toprule
		$i$ & ACC-2 ($\uparrow$) & F1 ($\uparrow$) & ACC-3 ($\uparrow$) & MAE ($\downarrow$) & Corr ($\uparrow$) \\ 
		\midrule
		1 & 38.5 & 38.5 & 31.9 & 0.218 & 0.287 \\
		2 & 39.0 & 38.8 & 31.6 & 0.219 & 0.287 \\
		3 & 38.7 & 38.7 & \textbf{32.1} & 0.216 & 0.290 \\
		4 & \textbf{39.3} & \textbf{39.2} & 32.0 & 0.216 & \textbf{0.294} \\
		5 & 38.6 & 38.6 & 32.1 & \textbf{0.215} & 0.292 \\
		\bottomrule
	\end{tabular}
\end{table}

\begin{table}[b] 
	\small
	\setlength{\tabcolsep}{2pt} 
	\renewcommand{\arraystretch}{1}
	\caption{ Remove the effect of transformer fusion module (w/o TF). }
	\label{tab10}
	\begin{tabular}{llllllll}
		\hline
		Models               & Dataset & Setting     & Metric      &             &         &       &       \\  \cmidrule(r){4-8} 
		&          & &ACC-5/3 & MAE   & Corr  \\ \hline
		w/o TF & MOSI    & complete  & 50.63$\mathrm{\left ( \Delta \downarrow  \right ) } $   & 0.741$\mathrm{\left ( \Delta \downarrow  \right ) } $ & 0.786$\mathrm{\left ( \Delta \downarrow  \right ) } $ \\
		&         & incomplete & 33.80$\mathrm{\left ( \Delta \downarrow  \right ) } $    & 0.114$\mathrm{\left ( \Delta \downarrow  \right ) } $ & 0.483$\mathrm{\left ( \Delta \uparrow  \right ) } $ \\
		& SIMS    & complete  & 65.20$\mathrm{\left ( \Delta \uparrow  \right ) } $    & 0.419$\mathrm{\left ( \Delta \downarrow  \right ) } $ & 0.602$\mathrm{\left ( \Delta \downarrow  \right ) } $ \\
		&         & incomplete & 31.60$\mathrm{\left ( \Delta \downarrow  \right ) } $    & 0.219$\mathrm{\left ( \Delta \downarrow  \right ) } $ & 0.282$\mathrm{\left ( \Delta \downarrow  \right ) } $ \\ \hline
	\end{tabular}
\end{table}

\begin{table}[]
	\normalsize
	\renewcommand{\arraystretch}{0.8}
	\setlength{\tabcolsep}{2pt} 
	\small
	\centering
	\caption{ Ablation of our proposed loss function on the CH-SIMS dataset.}
	\label{tab7}
	\begin{tabular}{llllllll}
		\hline
		& Loss  &   & ACC-2($\uparrow$) & F1($\uparrow$) & ACC-3($\uparrow$) & MAE($\downarrow$) & Corr($\uparrow$)   \\  \cmidrule(r){1-3} 
		dis                       & rec                       & sim                  &                   &                &                   &                   &                \\ \hline
		&                           &                           & 38.5              & 38.6           & 32.0              & 0.217             & 0.284          \\
		\checkmark &                           &                           & 38.3              & 38.1           & 31.0              & 0.227             & 0.279          \\
		& \checkmark &                           & 38.2              & 38.2           & 31.7              & 0.221             & 0.276          \\
		&                           & \checkmark & 38.4              & 38.4           & 31.8              & 0.217             & 0.283          \\
		\checkmark & \checkmark &                           & 38.4              & 38.3           & 31.0              & 0.225             & 0.283          \\
		& \checkmark & \checkmark & 38.7              & 38.7           & 32.0    & \textbf{0.215}    & 0.285          \\
		\checkmark &                           & \checkmark & 39.1              & 39.1           & 31.7              & 0.217             & 0.293          \\
		\checkmark & \checkmark & \checkmark & \textbf{39.3}     & \textbf{39.2}  & \textbf{32.0}     & 0.216             & \textbf{0.294} \\ \hline
	\end{tabular}
\end{table}

\begin{figure}[t]
	\centering
	\includegraphics[width=9cm,height=4cm]{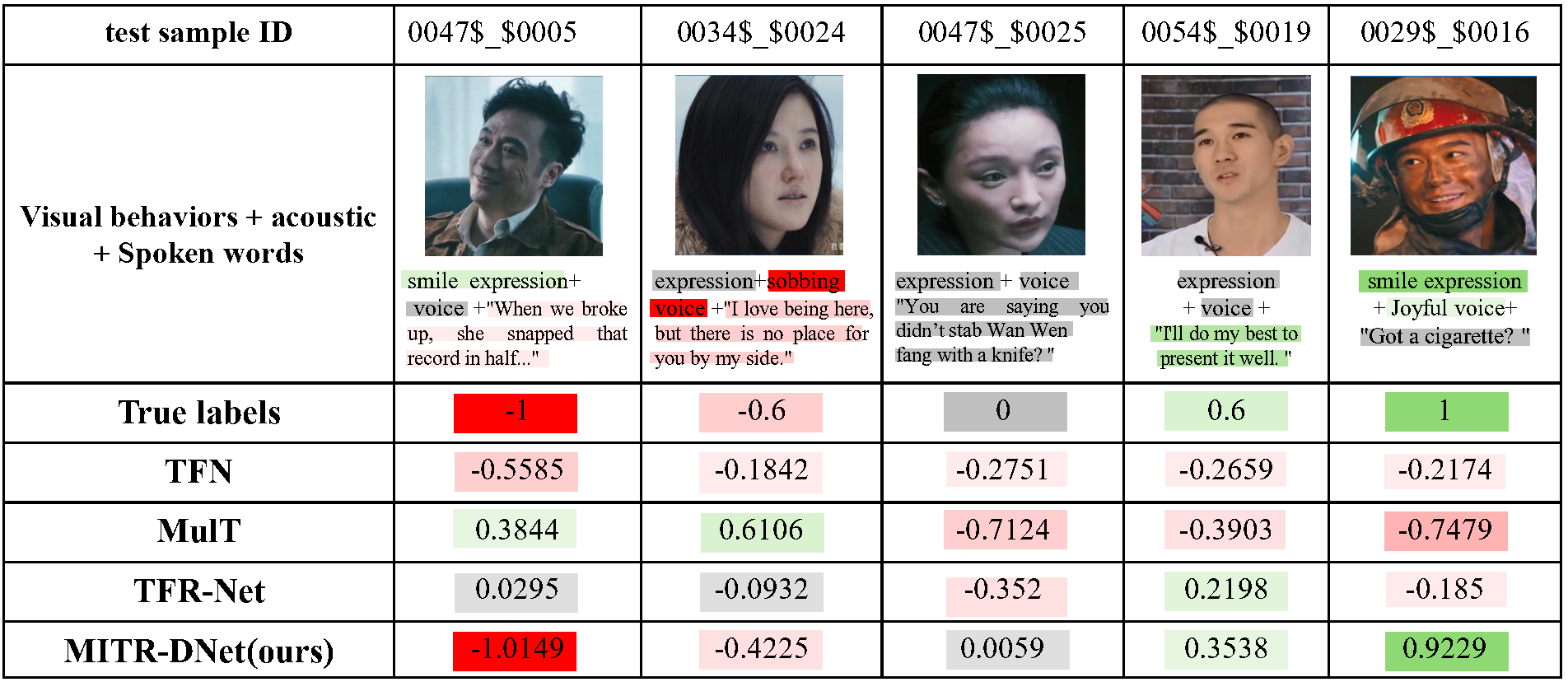} 
	\caption{Visualisation of model prediction labels on CH-SIMS dataset. Dark red indicates the most negative emotion (-1). Dark green indicates the most positive emotion (1). Dark grey indicates the most neutral emotion (0). The deeper the color, the stronger the emotion. }
	\label{33}
\end{figure}
\section{Experimental results}
\subsection{Comparison with state-of-the-art technology}\label{32}
Table~\ref{tab1} presents the performance of our model compared to current state-of-the-art models. Other models were reproduced using open-source code and the same experimental setup.  According to \cite{17}, the missing rates for the CMU-MOSI dataset range from 0.1 to 1.0, and for the CH-SIMS dataset, they range from 0.1 to 0.5. The final experimental results are calculated based on the Area Under Indicators Line Chart (AUILC)\cite{18}, which evaluates the model's overall performance at different missing rates. Several existing MSA models were selected for performance evaluation for the experimental comparison. TFN and MuIT models primarily target complete modality scenarios, while TFR-Net and EMT-DLFR are specifically designed to address the challenge of missing modalities. The experimental results indicate that MITR-DNet outperforms TFN and MuIT models in the complete modality setting and surpasses models designed for missing modalities in the intact modality setting on both datasets. Furthermore, in the incomplete modality setting, MITR-DNet achieves optimal or near-optimal scores across several metrics, demonstrating its effectiveness in handling the missing modality problem. Fig.~\ref{33} illustrates an example of label prediction, and MITR-DNet makes predictions closer to the ground truth label values by combining the three modalities, showing the effectiveness of our proposed MITR-DNet model in mitigating modal heterogeneity.
\begin{figure}[t]
	\centering
	\includegraphics[width=9cm]{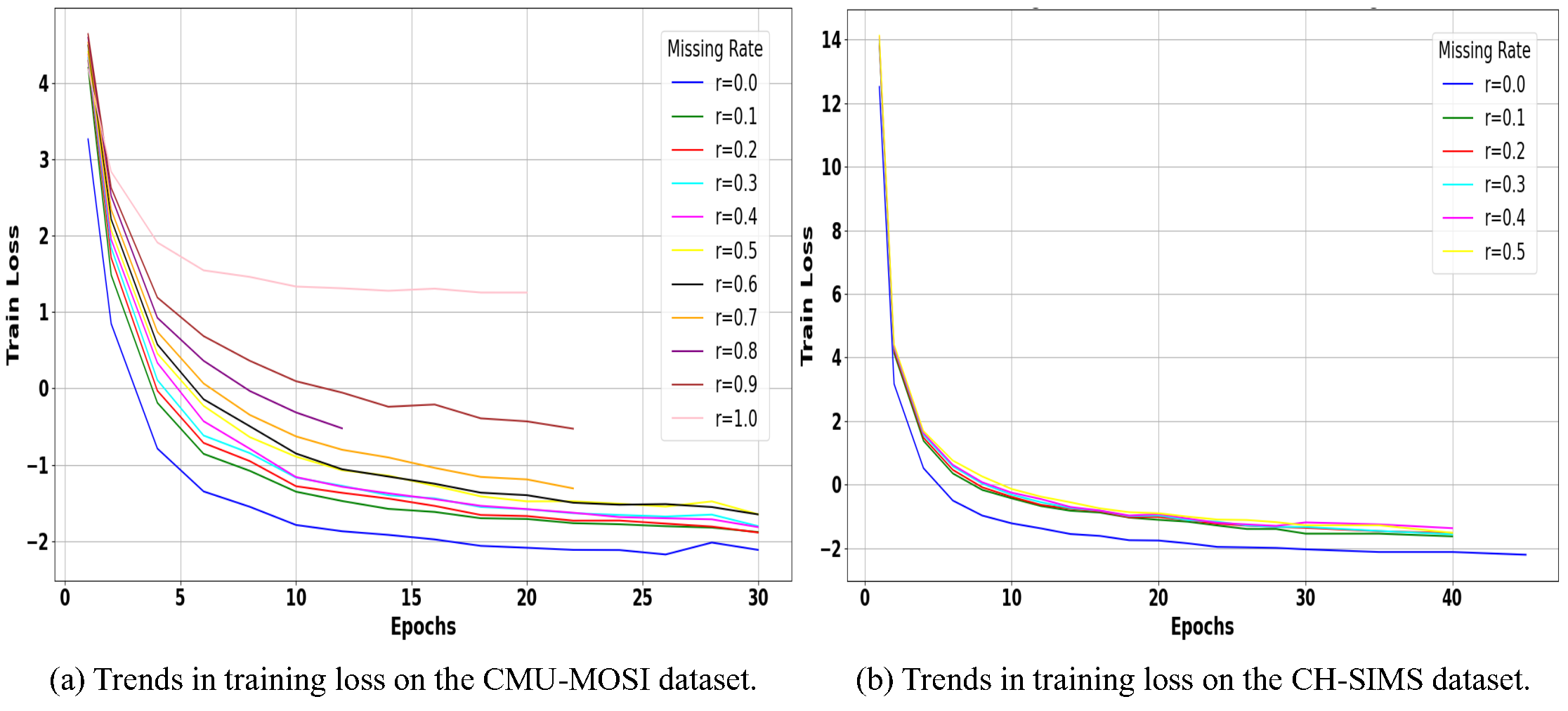} 
	\caption{Training loss trends under different modality missing rates}
	\label{14}
\end{figure}
\subsection{Ablation experiment}\label{4}
In the incomplete modality setup, we conducted ablation experiments to determine the optimal number of module $\mathcal{F}$ configurations, as shown in {Table~\ref{22}}. The results show that when $i=4$, the model exhibits optimal overall performance. Therefore, all our subsequent experiments were conducted using the $i=4$ configuration. 
{Table~\ref{tab10}} shows the results of the ablation experiments with the TF module removed. The results show a decrease in all performance metrics except for an increase in correlation (Corr) in the incomplete modality setting for the MOSI dataset and an increase in accuracy (ACC-3) in the complete modality setting for the SIMS dataset. These results emphasize the effectiveness of our design of the TF module. {Table~\ref{tab7}} highlights the importance of different loss function combinations under incomplete modality settings. The results show that our proposed MITR-DNet, which integrates distillation loss, reconstruction loss, and SimSiam loss, performs the best. Fig.~\ref{14} illustrates the training loss trends across various missing rates. The decreasing training loss throughout the process indicates that our proposed loss functions effectively handle MSA tasks with missing data.
\section{Conclusion}
This paper proposes a modality-invariant bidirectional temporal representation distillation network for missing multimodal sentiment analysis, effectively mitigating the issue of modality heterogeneity. However, data imbalance remains challenging due to the dataset's uneven distribution of emotional category samples. Mitigating the impact of data imbalance on multimodal sentiment analysis can be an important research direction.
\section*{Acknowledgment}
This study was funded by the Excellence Program Project of Tianshan, Xinjiang Uygur Autonomous Region, China (grant number 2022TSYCLJ0036), the Central Government Guides Local Science and Technology Development Fund Projects (grant number ZYYD2022C19), and the National Natural Science Foundation of China (grant numbers 62472368, 62463029, and 62303259).


\begin{thebibliography}{00}
\bibitem{1} M. Soleymani, D. Garcia, B. Jou, B. Schuller, S.-F. Chang, and M. Pantic, ``A survey of multimodal sentiment analysis,'' Image and Vision Computing, vol. 65, pp. 3--14, 2017, Elsevier.

\bibitem{2} A. Gandhi, K. Adhvaryu, S. Poria, E. Cambria, and A. Hussain, ``Multimodal sentiment analysis: A systematic review of history, datasets, multimodal fusion methods, applications, challenges and future directions,'' Information Fusion, vol. 91, pp. 424--444, 2023, Elsevier.

\bibitem{3} Y. Yu, Z. Jia, F. Shi, M. Zhu, W. Wang, and X. Li, ``WeaveNet: End-to-end audiovisual sentiment analysis,'' in International Conference on Cognitive Systems and Signal Processing, Springer, 2021, pp. 3--16.


\bibitem{41} Z. Qiu, M. Fu, Y. Yu, L. Yin, F. Sun, and H. Huang, ``Srtnet: Time domain speech enhancement via stochastic refinement,'' ICASSP 2023-2023 IEEE International Conference on Acoustics, Speech and Signal Processing (ICASSP), pp. 1--5, 2023.

\bibitem{42} Y. Yu, W. Huang, F. Sun, C. Chen, Y. Wang, and X. Liu, ``Sound adversarial audio-visual navigation,'' arXiv preprint arXiv:2202.10910, 2022.

\bibitem{43} Y. Yu, C. Chen, L. Cao, F. Yang, and F. Sun, ``Measuring acoustics with collaborative multiple agents,'' arXiv preprint arXiv:2310.05368, 2023.

\bibitem{44} Y. Yu, L. Cao, F. Sun, C. Yang, H. Lai, and W. Huang, ``Echo-enhanced embodied visual navigation,'' Neural Computation, vol. 35, no. 5, pp. 958--976, 2023, MIT Press.

\bibitem{45} Y. Yu, L. Cao, F. Sun, X. Liu, and L. Wang, ``Pay self-attention to audio-visual navigation,'' in Proceedings of the British Machine Vision Conference (BMVC), 2022, p. 46.
\bibitem{46} A. H. Yazdavar, M. S. Mahdavinejad, G. Bajaj, W. Romine, A. Sheth, A. H. Monadjemi, K. Thirunarayan, J. M. Meddar, A. Myers, J. Pathak, et al., ``Multimodal mental health analysis in social media,'' PLOS ONE, vol. 15, no. 4, p. e0226248, 2020, Public Library of Science.
\bibitem{4} X. Jiao, L. Wang, and Y. Yu, ``MFHCA: Enhancing speech emotion recognition via multi-spatial fusion and hierarchical cooperative attention,'' arXiv preprint arXiv:2404.13509, 2024.
\bibitem{5} X. Wang, L. Wang, Y. Yu, and X. Jiao, ``PCQ: Emotion recognition in speech via progressive channel querying,'' in International Conference on Intelligent Computing, Springer, 2024, pp. 264--275.

\bibitem{6} A. Zadeh, M. Chen, S. Poria, E. Cambria, and L.-P. Morency, ``Tensor fusion network for multimodal sentiment analysis,'' arXiv preprint arXiv:1707.07250, 2017.

\bibitem{7} Y.-H. H. Tsai, S. Bai, P. P. Liang, J. Z. Kolter, L.-P. Morency, and R. Salakhutdinov, ``Multimodal transformer for unaligned multimodal language sequences,'' in Proceedings of the Conference. Association for Computational Linguistics. Meeting, vol. 2019, NIH Public Access, 2019, p. 6558.

\bibitem{8} D. Hazarika, R. Zimmermann, and S. Poria, ``MISA: Modality-invariant and-specific representations for multimodal sentiment analysis,'' in Proceedings of the 28th ACM International Conference on Multimedia, 2020, pp. 1122--1131.

\bibitem{9} W. Yu, H. Xu, Z. Yuan, and J. Wu, ``Learning modality-specific representations with self-supervised multi-task learning for multimodal sentiment analysis,'' in Proceedings of the AAAI Conference on Artificial Intelligence, vol. 35, no. 12, 2021, pp. 10790--10797.

\bibitem{10} R. Huan, G. Zhong, P. Chen, and R. Liang, ``Unimf: A unified multimodal framework for multimodal sentiment analysis in missing modalities and unaligned multimodal sequences,'' IEEE Transactions on Multimedia, 2023, IEEE.

\bibitem{11} R. Guerrero, H. X. Pham, and V. Pavlovic, ``Cross-modal retrieval and synthesis (X-MRS): Closing the modality gap in shared subspace learning,'' in Proceedings of the 29th ACM International Conference on Multimedia, 2021, pp. 3192--3201.

\bibitem{12} D. Jia, A. Hermans, and B. Leibe, ``Domain and modality gaps for lidar-based person detection on mobile robots,'' arXiv preprint, 2021.


\bibitem{13} D. Hazarika, R. Zimmermann, and S. Poria, ``MISA: Modality-invariant and-specific representations for multimodal sentiment analysis,'' in Proceedings of the 28th ACM International Conference on Multimedia, 2020, pp. 1122--1131.

\bibitem{14} Y. Li, Y. Wang, and Z. Cui, ``Decoupled multimodal distilling for emotion recognition,'' in Proceedings of the IEEE/CVF Conference on Computer Vision and Pattern Recognition, 2023, pp. 6631--6640.

\bibitem{15} D. Yang, S. Huang, H. Kuang, Y. Du, and L. Zhang, ``Disentangled representation learning for multimodal emotion recognition,'' in Proceedings of the 30th ACM International Conference on Multimedia, 2022, pp. 1642--1651.

\bibitem{16} J. Li, S. Cai, L. Li, R. Sun, G. Yuan, and R. Zhu, ``MIT-FRNet: Modality-invariant temporal representation learning-based feature reconstruction network for missing modalities,'' Expert Systems with Applications, vol. 249, p. 123655, 2024, Elsevier.

\bibitem{17} L. Sun, Z. Lian, B. Liu, and J. Tao, ``Efficient multimodal transformer with dual-level feature restoration for robust multimodal sentiment analysis,'' IEEE Transactions on Affective Computing, vol. 15, no. 1, pp. 309--325, 2023, IEEE.

\bibitem{18} Z. Yuan, W. Li, H. Xu, and W. Yu, ``Transformer-based feature reconstruction network for robust multimodal sentiment analysis,'' in Proceedings of the 29th ACM International Conference on Multimedia, 2021, pp. 4400--4407.


\bibitem{20} A. van den Oord, S. Dieleman, H. Zen, K. Simonyan, O. Vinyals, A. Graves, N. Kalchbrenner, A. Senior, and K. Kavukcuoglu, ``Wavenet: A generative model for raw audio,'' arXiv preprint arXiv:1609.03499, 2016.


\bibitem{21} C. Zhu, M. Chen, S. Zhang, C. Sun, H. Liang, Y. Liu, and J. Chen, ``SKEAFN: Sentiment knowledge enhanced attention fusion network for multimodal sentiment analysis,'' Information Fusion, vol. 100, p. 101958, 2023, Elsevier.

\bibitem{22} Y. Wu, Z. Lin, Y. Zhao, B. Qin, and L.-N. Zhu, ``A text-centered shared-private framework via cross-modal prediction for multimodal sentiment analysis,'' in Findings of the Association for Computational Linguistics: ACL-IJCNLP 2021, 2021, pp. 4730--4738.

\bibitem{23} X. Chen and K. He, ``Exploring simple siamese representation learning,'' in Proceedings of the IEEE/CVF Conference on Computer Vision and Pattern Recognition, 2021, pp. 15750--15758.

\bibitem{24} A. Zadeh, R. Zellers, E. Pincus, and L.-P. Morency, ``MOSI: Multimodal corpus of sentiment intensity and subjectivity analysis in online opinion videos,'' arXiv preprint arXiv:1606.06259, 2016.

\bibitem{25} W. Yu, H. Xu, F. Meng, Y. Zhu, Y. Ma, J. Wu, J. Zou, and K. Yang, ``CH-SIMS: A Chinese multimodal sentiment analysis dataset with fine-grained annotation of modality,'' in Proceedings of the 58th Annual Meeting of the Association for Computational Linguistics, 2020, pp. 3718--3727.


\end{thebibliography}
\end{document}